\documentclass[11pt,a4paper]{article}
\usepackage[hyperref]{acl2021}
\usepackage{times}
\usepackage{latexsym}

\usepackage{microtype}
\usepackage{amsfonts}
\usepackage{amsmath}
\usepackage{amssymb}
\usepackage{graphicx}
\usepackage{xspace}
\usepackage{todonotes}
\usepackage{fancyvrb}

\newcommand\mname{\textsc{Turing}\xspace}

\newcommand{\ench}{\mathcal{H}}

\newcommand{\secvspace}{0em}

\aclfinalcopy 

\title{\textsc{Turing}: an Accurate and Interpretable
  Multi-Hypothesis Cross-Domain \\
  Natural Language Database Interface
  }

\author{Peng Xu\thanks{~ Equal contribution},~~
  Wenjie Zi$^*$,~~
  Hamidreza Shahidi,~~
{\'A}kos K{\'a}d{\'a}r,~~ Keyi Tang, \\
  \bf Wei Yang,~~ Jawad Ateeq,~~ Harsh Barot,~~ Meidan Alon,~~ Yanshuai Cao \\  
  Borealis AI \\
  \scriptsize \texttt{\{peng.z.xu, wenjie.zi, hamidreza.shahidi, akos.kadar, keyi.tang\}@borealisai.com}\\
  \scriptsize \texttt{\{wei.yang, jawad.ateeq, harsh.barot, meidan.alon, yanshuai.cao\}@borealisai.com}
}

\date{}

\begin{document}
\maketitle
\begin{abstract}
  A natural language database interface (NLDB) can democratize data-driven insights for non-technical users. However, existing Text-to-SQL semantic parsers cannot achieve high enough accuracy in the cross-database setting to allow good usability in practice. This work presents \mname \footnote{System demo at \url{https://turing.borealisai.com/}; video at \url{https://vimeo.com/537429187/9a5d41f446}}, a NLDB system toward bridging this gap.
The cross-domain semantic parser of \mname with our novel value prediction method achieves $75.1\%$ execution accuracy, and $78.3\%$ top-5 beam execution accuracy on the Spider validation set \cite{yu2018spider}.
To benefit from the higher beam accuracy, we design an interactive system where the SQL hypotheses in the beam are explained step-by-step in natural language, with their differences highlighted. The user can then compare and judge the hypotheses to select which one reflects their intention if any. The English explanations of SQL queries in \mname are produced by our high-precision natural language generation system based on synchronous grammars.

\end{abstract}

\section{Introduction}
\vspace{\secvspace}
Today a vast amount of knowledge is hidden in structured datasets, not directly accessible to non-technical users who are not familiar with the corresponding database query language like SQL or SPARQL. Natural language database interfaces (NLDB) enable everyday users to interact with databases \cite{zelle1996learning,popescu2003towards,Li2014NaLIRAI,zeng-etal-2020-photon}. However, correctly translating natural language to executable queries is challenging, as it requires resolving all the ambiguities and subtleties of natural utterances for precise mapping. Furthermore, quick deployment and adoption for NLDB require zero-shot transfer to new databases without an in-domain text-to-SQL parallel corpus, \emph{i.e.\ } cross-database semantic parsing (SP), making the translation accuracy even lower. Finally, unlike in other NLP applications where partially correct results can still provide partial utility, a SQL query with a slight mistake could cause negative utility if trusted blindly or confusing to users. 

The recent Spider benchmark \cite{yu-etal-2018-spider} captures this cross-domain problem, and the state-of-the-art methods merely achieve around $70\%$ execution accuracy at the time of this submission \footnote{\url{https://yale-lily.github.io/spider}}. Meanwhile, generalization to datasets collected under different protocols is even weaker \cite{suhr-etal-2020-exploring}. Finally, users generally have no way to know if the NLDB made a mistake except in very obvious cases. The high error rate combined with the overall system's opacity makes it hard for users to trust any output from the NLDB.

Our key observation is that our model's top-5 accuracy on Spider is $78.3\%$, significantly higher than the previous best single-model method at around $68\%$, and our own top-1 accuracy. Top-5 accuracy is the proportion of times when one of the top five hypotheses from beam-search inference is correct (in execution accuracy evaluation). For top-5 accuracy to be relevant in practice, a non-technical user needs to be able to pick the correct hypothesis from the candidate list.
To this end, we design a feedback system that can unambiguously explain the top beam-search results while presenting the differences intuitively and visually.  Users can then judge which, if any, of the parses correctly reflects their intentions. The explanation system uses a hybrid of two synchronous context-free grammars, one shallow and one deep. Together, they achieve good readability for the most frequent query patterns while near-complete coverage overall.

Our system, \mname, is not only interpretable, but also a highly accurate cross-domain NLDB.
Our semantic parser is based on the one in \citet{Xu2020OptimizingDT}, which does not handle value prediction like many other previous state-of-the-art models on Spider. Compared to previous executable semantic parsers, we achieve significant gains with a number of techniques, but predominantly by drastically simplifying the learning problem in value prediction. The model only needs to identify the text span providing evidence for the ground-truth value. The noisy long tail text normalization step required for producing the actual value is offloaded to a deterministic search phase in post-processing. 

In summary, this work presents two steps towards a more robust NLDB:
\begin{enumerate}
    \item A state-of-the-art text-to-SQL parsing system with the best top-1 execution accuracy on the Spider development set.
    \item A way to relax usability requirement from top-1 accuracy to top-k accuracy by explaining the different hypotheses in natural language with visual aids.
\end{enumerate}

\section{System Overview}
\begin{figure*}[ht!]
  \centering
  \includegraphics[width=.98\linewidth]{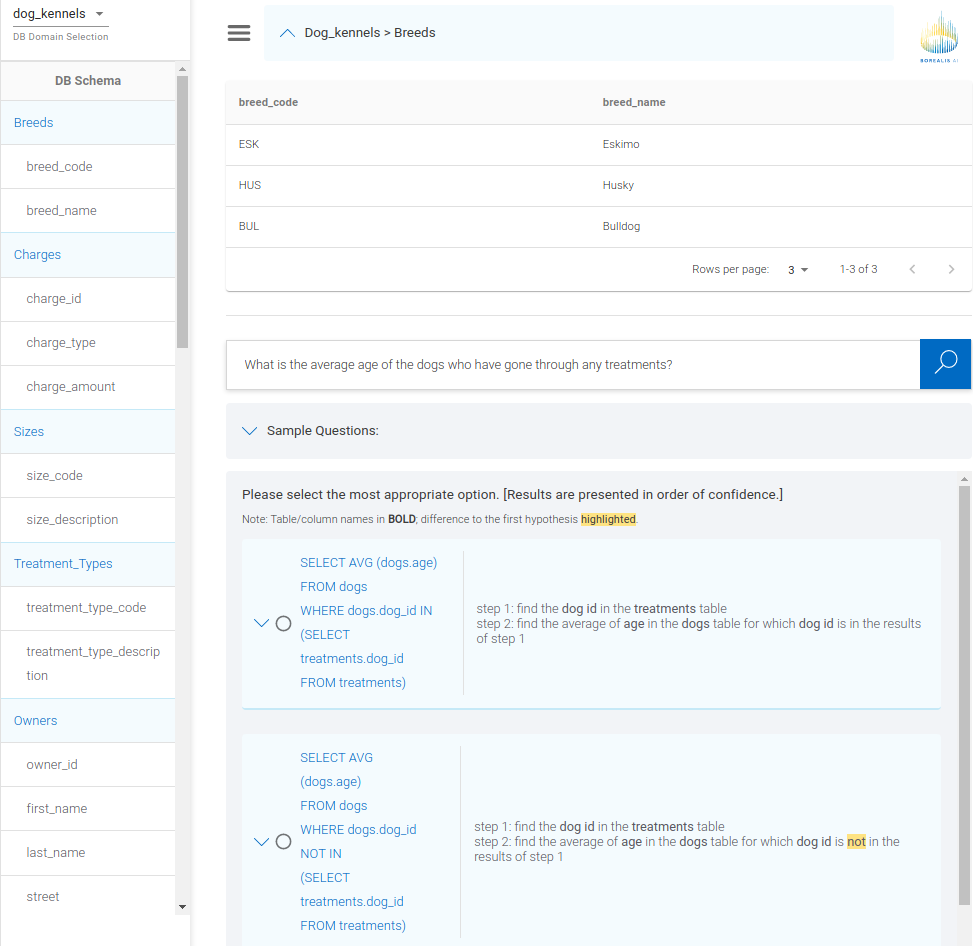}
  \caption{\mname system in action: the user selected database ``Dog\_kennels''; the left and top panels show the database schema and table content. The user then entered ``What is the average age of the dogs who have gone through any treatments?'' in the search box. This question is run through the semantic parser producing multiple SQL hypotheses from beam-search, which are then explained step-by-step as shown. The differences across the hypotheses are highlighted. The tokens corresponding to table and columns are in bold. If there were more valid hypotheses, a ``Show more'' button would appear to reveal the additional ones.} \label{fig:interface}
\end{figure*}



As shown in Figure \ref{fig:interface}, \mname's interface has two main components: the database browser showing schema and selected database content, and the search panel where the users interact with the parser. Figure \ref{fig:interface} caption describes the typical user interaction using an example.

Behind the front-end interface, \mname consists of an executable cross-domain semantic parser trained on Spider that maps user utterances to SQL query hypotheses, the SQL execution engine that runs the queries to obtain answers, and the explanation generation module that produces the explanation text and the meta-data powering explanation highlighting. The next sections will describe the semantic parsing and explanation modules.

\section{Semantic Parser}
\label{sec:sp}
\vspace{\secvspace}
The backbone of \mname is a neural semantic parser which generates an executable SQL query $T$ given a user question $Q$ and the database schema $\mathcal{S}$.
We follow the state-of-the-art system \cite{Xu2020OptimizingDT}, but extend it to generate executable SQL query instead of ignoring values in the SQL query, like many other top systems \cite{wang2019rat, guo2019towards} on the Spider leaderboard.

On the high-level, our SP adopts the grammar-based framework following TranX \cite{yin2018tranx} with an encoder-decoder neural architecture.
A grammar-based transition system is designed to turn the generation process of the SQL abstract syntax tree (AST) into a sequence of tree-constructing actions to be predicted by the parser.
The encoder $f_{\text{enc}}$ jointly encodes both the user question $Q=q_1\dots q_{\lvert Q \rvert}$ and database schema $\mathcal{S}=\{s_1,\dots, s_{\lvert \mathcal{S} \rvert}\}$ consisting of tables and columns in the database. 
The decoder $f_{\text{dec}}$ is a transition-based abstract syntax decoder, which uses the encoded representation $\ench$ to predict the target SQL query $T$.
The decoder also relies on the transition system to convert the AST constructed by the predicted action sequences to the executable surface SQL query.

To alleviate unnecessary burden on the decoder, we introduce two novel modifications to the transition system to handle the schema and value decoding.
With simple, but effective value-handling, inference and regularization techniques applied on this transition system, we are able to push the execution accuracy much higher for better usability.

\subsection{Transition System}
\vspace{\secvspace}
\label{sec:ts}
Our transition system has four types of action to generate the AST, including (1) {\bf ApplyRule}[$r$] which applies a production rule $r$ to the latest generated node in the AST; (2) {\bf Reduce} which completes the generation of the current node; (3) {\bf SelectColumn}[$c$] which chooses a column $c$ from the database schema $\mathcal{S}$; (4) {\bf CopyToken}[$i$] which copies a token $q_i$ from the user question $Q$.

There are two key distinctions of our transition system with the previous systems.
First, our transition system omits the action type {\bf SelectTable} used by other transition-based SP systems \cite{wang2019rat,guo2019towards}. This is made possible by attaching the corresponding table to each column, so that the tables in the target SQL query can be deterministically inferred from the predicted columns.
Second, we simplify the value prediction by always trying to copy from the user question, instead of  applying the {\bf GenToken}[$v$] action \cite{yin2018tranx} which generates tokens from a large vocabulary or choose from a pre-processed picklist \cite{lin2020bridging}.
Both of the changes constrain the output space of the decoder to ease the learning process, but the latter change unrealistically assumes that the values are always explicitly mentioned in the question.
To retain the generation flexibility without putting excessive burden on the decoder, we propose a conceptually simple but effective strategy to handle the values next.

\subsection{Handling Values}
\vspace{\secvspace}
Value prediction is a challenging, but crucial component of NLDBs, however, only limited efforts are committed to handling values properly in the current cross-domain SP literature.
Value mentions are usually noisy, if mentioned explicitly at all, requiring commonsense or domain knowledge to be inferred.
On the other hand, the number of possible values in a database can be huge, leading to sparse learning signals if the model tries to choose from the possible value candidates.

Instead of attempting to predict the actual values directly, our SP simply learns to identify the input text spans providing evidence for the values.
As mentioned earlier, we introduce the {\bf CopyToken} action to copy an input span from the user question, indicating the clues for this value.
The ground-truth {\bf CopyToken}[$i$] actions are obtained from a tagging strategy based on heuristics and fuzzy string matching between the user question and the gold values.
As a result, the decoder is able to focus on understanding the question without considering other complexities of the actual values which are difficult to learn.
If the values are only implicitly mentioned in the user question, nothing is copied from the user question.
We leave the identification of the actual values to a deterministic search-based inference in post-processing, after the decoding process. This yields a simpler learning task as the neural network does not need to perform domain-specific text normalization such as mapping ``female'' to ``F'' for some databases. 

Given the schema, the predicted SQL AST and the database content, the post-processing first identifies the corresponding column type (\emph{number, text, time}), operation type (\emph{like, between, $>$, $<$, =, ...}), and aggregation type (\emph{count, max, sum, ...}). Based on these types, it infers the type and normalization required for the value. If needed, it then performs fuzzy-search in the corresponding column's values in the database.
When nothing is copied, a default value is chosen based on some heuristics (\emph{e.g.}, when there exist only two element ``Yes" and ``No" in the column, the default value is ``Yes"); otherwise, the most frequent element in the column is chosen.
Searching the database content can also be restricted to a picklist for privacy reasons like previous works \cite{zeng-etal-2020-photon,lin2020bridging}.

Another benefit of this simple value handling strategy is the ease to explain.
The details are presented in the Sec.~\ref{sec:explanation}.

\subsection{Encoder-Decoder}
\label{sec:encoder}
\vspace{\secvspace}
Our encoder architecture follows \citet{Xu2020OptimizingDT}. 
The encoder, $f_{\text{enc}}$, maps the user question $Q$ and the schema $\mathcal{S}$ to a joint representation $\ench = \{\phi^q_1, \ldots,\phi^q_{\lvert Q \rvert} \} \cup \{\phi^s_1, \ldots,\phi^s_{\lvert \mathcal{S} \rvert} \}$. It contextualizes the question and schema jointly through both the RoBERTA-Large model similar to \cite{guo2019towards}, as well as through the additional sequence of $24$ relation-aware transformer (RAT) \cite{wang2019rat} layers. 
As mentioned in Section~\ref{sec:ts}, tables are not predicted directly but inferred from the columns, so we augment the column representations by adding the corresponding table representations after the encoding process.

We use a LSTM decoder $f_{\text{dec}}$ to generate the action sequence $A$. 
Formally, the generation process can be formulated as $\Pr(A|\ench) = \prod_t \Pr(a_t|a_{<t}, \ench)$ where $\ench$ is the encoded representations outputted by the encoder $f_{\text{enc}}$.
The LSTM state is updated following \citet{wang2019rat}: $\pmb m_t, \pmb h_t=f_{\text{LSTM}}([\pmb a_{t-1} \Vert \pmb z_{t-1} \Vert \pmb h_{p_t} \Vert \pmb a_{p_t} \Vert \pmb n_{p_t}], \pmb m_{t-1}, \pmb h_{t-1})$, where $\pmb m_t$ is the LSTM cell state, $\pmb h_t$ is the LSTM output at step $t$, $\pmb a_{t-1}$ is the action embedding of the previous step, $\pmb z_{t-1}$ is the context representation computed using multi-head cross-attention of $\pmb h_{t-1}$ over $\ench$, $p_t$ is the step corresponding to the parent AST node of the current node, and $\pmb n$ is the node type embedding.
For {\bf ApplyRule}[$r$], we compute $\Pr(a_t=\textbf{ApplyRule}[r]|a_{<t}, \ench)=\text{softmax}_r(g(\pmb z_t))$ where $g(\cdot)$ is a 2-layer MLP.
For {\bf SelectColumn}[$c$], we use the memory augmented pointer network following \citet{guo2019towards}.
For {\bf CopyToken}[$i$], a pointer network is employed to copy tokens from the user question $Q$ with a special token indicating the termination of copy.

\subsection{Column Label Smoothing}
\label{sec:reg}
\vspace{\secvspace}
One of the core challenges for cross-domain SP is to generalize to unseen domains without overfitting to some specific domains during training.
Empirically, we observe that applying uniform label smoothing \cite{szegedy2016rethinking} on the objective term for predicting {\bf SelectColumn}[$c$] can effectively address the overfitting problem in the cross-domain setting.
Formally, the cross-entropy for a ground-truth column $c^*$ we optimize becomes $(1-\epsilon) * \log p(c^*) + \frac{\epsilon}{K} * \sum_c\log p(c)$, 
where $K$ is the number of columns in the schema, $\epsilon$ is the weight of the label smoothing term, and $p(\cdot) \triangleq \Pr(a_t=\textbf{SelectColumn}[\cdot]|a_{<t}, \ench)$.

\subsection{Weighted Beam Search}
\vspace{\secvspace}
During inference, we use beam search to find the high-probability action sequences.
As mentioned above, column prediction is prone to overfitting in the cross-domain setting. 
In addition, value prediction is dependent on the column prediction, that is, if a column is predicted incorrectly, the associated value has no chance to be predicted correctly.
As a result, we introduce two hyperparameters controlling influence based on the action types in the beam, with a larger weight $\alpha>1$ for {\bf SelectColumn} and a smaller weight $0<\beta<1$ for {\bf CopyToken}.

\section{Explanation Generation}
\label{sec:explanation}
\vspace{\secvspace}
The goal of the explanation generation system is to unambiguously describe what the semantic parser understands as the user's command and allow the user to easily interpret the differences across the multiple hypotheses. Therefore, unlike a typical dialogue system setting where language generation diversity is essential, controllability and consistency are of primary importance.
The generation not only needs to be 100\% factually correct, but the differences in explanation also need to reflect the differences in the predicted SQLs, no more and no less. Therefore, we use a deterministic rule-based generation system instead of a neural model.

Our explanation generator is a hybrid of two synchronous context-free grammar (SCFG) systems combined with additional heuristic post-processing steps. The two grammars trade off readability and coverage. One SCFG is shallow and simple, covering the most frequent SQL queries; the other is deep and more compositional, covering the tail of query distribution that our SP can produce for completeness. The SCFG can produce SQL and English explanation parallel. Given a SQL query, we parse it under the grammar to obtain a derivation, which we then follow to obtain the explanation text. At inference time, for a given question, if any of the SQL hypotheses cannot be parsed using the shallow SCFG, then we move onto the deep one.

\subsection{Details of the Grammars}
Using the deep SQL syntax trees allows almost complete coverage on the Spider domains. However, these explanations can be unnecessarily verbose as the generation process faithfully follows the re-ordered AST without 1.) compressing repeated mentions of schema elements when possible 2.) summarizing tedious details of the SQL query into higher level logical concepts. Even though these explanations are technically correct, practical explanation should allow users to spot the difference between queries easily. To this end, we design the shallow grammar similarly to the template-based explanation system in \citet{elgohary2020speak}, which simplifies the SQL parse trees by collapsing large subtrees into a single tree fragment. In the resulting shallow parses production rules yield non-terminal nodes corresponding to 1.) anonymized SQL templates 2.) UNION, INTERSECT, or EXCEPT operations of two templates 3.) or a template pattern followed by ORDER-BY-LIMIT clause. Our shallow but wide grammar has $64$ rules with those non-terminal nodes. The pre-terminal nodes are place-holders in the anonymized SQL queries such as Table name, Column name, Aggregation operator and so on. Finally, the terminal nodes are the values filling in the place holders. The advantage of this grammar is that each high-level SQL template can be associated with an English explanation template that reveals the high level logic and abstracts away from the details in the concrete queries.
To further reduce the redundancy, we make assumptions to avoid unnecessarily repeating table and column names. Table.\ \ref{tab:rule} showcases some rules from the shallow SCFG and one example of explanation. In practice, around $75\%$ of the examples in the Spider validation set have all beam hypotheses from our SP model parsable by the shallow grammar, with the rest handled by the deep grammar. The deep grammar has less than $50$ rules. But because it is more compositional, it covers $100\%$ of the valid SQLs that can be generated by our semantic parser. Some sample explanation by the deep grammar can be found in Table.\ \ref{tab:deep_examples}.

Finally, whenever the final value in the query differs from original text span due to post-processing, a sentence in the explanation states the change explicitly for clarity. For example, ``{\itshape`Asian' in the question is matched to `Asia' which appears in the column Continent.}''

\begin{table*}[h]
\centering
\begin{tabular}{|c|}
\hline
\begin{minipage}{.95\linewidth}
\vspace{1ex}
\begin{Verbatim}[fontsize=\tiny]
S -> P
S -> P UNION P
P -> (SELECT <T_0>.<C_0> FROM <T_1> GROUP BY <T_2>.<C_1> HAVING <AOps_0> ( <T_3>.<C_2> ) <WOps_0> <L_0>,       
      find the different values of the {<C_0>} in the {<T_1>} whose {<AOps_0>} the {<C_2>} {<WOps_0>} {<L_0>})
\end{Verbatim}
\vspace{.1ex}
\end{minipage}\\
\hline
\begin{minipage}{.95\linewidth}
\vspace{1ex}
\begin{Verbatim}[fontsize=\scriptsize]
step 1: find the average of product price in the products table
step 2: find the different values of the product type code in the products table 
        whose average of the product price is greater than the results of step 1
\end{Verbatim}
\vspace{.1ex}
\end{minipage}\\
\hline
\end{tabular}
\caption{Sample shallow grammar production rules and one example explanation.}
\label{tab:rule}
\end{table*}

\begin{table*}[h]
\centering
\begin{tabular}{|c|}
\hline
\begin{minipage}{.95\linewidth}
\vspace{1ex}
\begin{Verbatim}[fontsize=\scriptsize]
Step 1: find the entries in the employee table whose age is less than 30.0.
Step 2: among these results, for each city of the employee table,
        where the number of records is more than 1, find city of the employee table.        
---------------
"30" in the question is converted to 30.
"one" in the question is converted to 1.
\end{Verbatim}
\vspace{.1ex}
\end{minipage}\\%
\hline
\begin{minipage}{.95\linewidth}
\vspace{1ex}
\begin{Verbatim}[fontsize=\scriptsize]
Step 1: find combinations of entries in the employee table, the hiring table and the shop table
        for which employee id of the employee table is equal to employee id of the hiring table
        and shop id of the hiring table is equal to shop id of the shop table.
Step 2: among these results, for each shop id of the shop table,
        find the average of age of the employee table and shop id of the shop table.
\end{Verbatim}
\vspace{.1ex}
\end{minipage}\\
\hline
\end{tabular}
\caption{Examples of explanation by the deep grammar. The first example also showcases the additional explanation for value post-processing.}
\label{tab:deep_examples}
\end{table*}


\section{Quantitative Evaluations}
\vspace{\secvspace}
\begin{table}[h]
\centering
\begin{tabular}{l c c}
\hline
\bf Model  & \bf Exec \\ \hline
{\scriptsize GAZP + BERT \cite{zhong2020grounded}}  & {\small $59.2$} \\ 
{\scriptsize Bridge v2 + BERT \cite{lin2020bridging}}  & {\small $68.0$} \\
{\scriptsize Bridge v2 + BERT (ensemble)}  & {\small $70.3$} \\  \hline
{\scriptsize Turing + RoBERTa} & {\small $\mathbf{75.1} (\text{best}), 73.8 \pm 0.7$}\\
\hline
\end{tabular}
\caption{{\bf Exec} accuracy on the Spider development set.}
\label{tab:spider}
\end{table}
\begin{table}[h]
\centering
\begin{tabular}{l c}
\hline
\bf Model & \bf Exec \\ \hline
Turing + RoBERTa & $73.8 \pm 0.7$ \\ \hline
\quad w/o. value post-processing & $67.2 \pm 0.8$ \\
\quad w/o. column label smoothing & $73.1 \pm 1.2$ \\
\quad w/o. weighted beam search & $73.5 \pm 0.7$\\ \hline
\quad top 3 in the beam & $77.3 \pm 0.4$\\
\quad top 5 in the beam & $78.3 \pm 0.3$\\
\hline
\end{tabular}
\caption{Ablation study on various techniques used in \mname. We use 5 runs with different random seeds.}
\label{tab:ablation}
\end{table}

\noindent\textbf{Implementation Details.}
We apply the DT-Fixup technique from \cite{Xu2020OptimizingDT} to train our semantic parser and mostly re-use their hyperparamters.
The weight of the column label smoothing term $\epsilon$ is $0.2$.
Inference uses a beam size of $5$ for the beam search.
We set the column weight as $\alpha=3$ and the value weight as $\beta=0.1$. 

\noindent\textbf{Dataset.} We use Spider \cite{yu2018spider}, a complex and cross-domain Text-to-SQL semantic parsing benchmark, which has $10,180$ questions, $5,693$ queries covering $200$ databases in $138$ domains.
All our experiments are evaluated based on the development set.
We use the execution match \emph{with} values ({\bf Exec}) evaluation metrics.

\noindent\textbf{Results on Spider.} We compare \mname with the top systems on the Spider execution leaderboard that have published reports with execution accuracy on the development set as well. 
As seen from Table~\ref{tab:spider}, our single model significantly outperforms the previous state of the art in terms of {\bf Exec} accuracy on the development set.

\noindent\textbf{Ablation Study.} Table~\ref{tab:ablation} shows an ablation study of various techniques in \mname.
We can see that removing the value post-processing decreases the accuracy significantly, showing that copying alone is not enough due to the mismatch in linguistic variation and the schema specific normalization.
The effectiveness of the proposed column label smoothing and weighted beam search are also reflected by the {\bf Exec} accuracy on Spider.
Furthermore, simply adding more hypotheses in the beam can significantly boost the coverage of the correct predictions, leading to $4.5\%$ accuracy gain over the top one accuracy.
By combining all these techniques together, \mname achieves an overall performance gain above $10\%$ over the previous best single model system ($68.0\%$ of Bridge v2). \footnote{\citet{rubin2020smbop} updated a version (April 11th 2021) around the time of this submission with a dev accuracy of $75\%$ (missing from the first version), and a test accuracy of $71.1\%$ significantly higher than the original $60.5\%$.}

\section{Related Work}
\vspace{\secvspace}
\paragraph{Executable Cross-database Semantic Parsing.}
Early NLDB systems use rule-based parsing \cite{zelle1996learning,Li2014NaLIRAI} and cannot handle the diversity of natural language in practice. Neural semantic parsing is more promising for coverage but is still brittle in real-world applications where queries can involve novel compositions of learned patterns \cite{finegan-dollak-etal-2018-improving,shaw2020compositional}. Furthermore, to allow plug-and-play on new databases, the underlying semantic parser may not be trained on in-domain parallel corpus but needs to transfer across domains in a zero-shot fashion.

Executable cross-database semantic parsing is even more challenging. Many of the previous work only tackle the cross-domain part, omitting the value prediction problem required for executable queries \cite{guo2019towards,wang2019rat,choi2020ryansql,Xu2020OptimizingDT}. Unlike the output space of predicting the SQL sketch or columns, the value prediction output space is much less constrained. The correct value depends on the source question, the SQL query, the type information of the corresponding column, as well as the database content. This complexity combined with limited training data in standard benchmark datasets like Spider makes the task very difficult. Some previous works directly learn to predict the values \cite{yin2018tranx,guo2020content} on WikiSQL \cite{zhong2017seq2sql}, but does not generalize in cross-domain settings. On Spider, \citet{zeng-etal-2020-photon} and \citet{lin2020bridging} build a candidate list of values first and learn a pointer network to select from the list. \mname instead learns a pointer network to identify the input source span that provides evidence for the value instead of directly the value as previously described. Identification of the actual value is offloaded to post-processing. From a system perspective, it is also simpler for a power user of the NLDB to upload a domain-specific term description/mapping which can extend the heuristic-search-based value post-processing instantly rather than relying on re-training.

\paragraph{Query Explanation.} Explaining structured query language has been studied in the past \cite{abs-0909-1786,5447824,NgomoBULG13,xu-etal-2018-sql}. Full NLDB systems can leverage explanations to correct mistakes with user feedback \cite{elgohary2020speak}, or to prevent mistakes by giving clarifications \cite{zeng-etal-2020-photon}. However, these methods can only handle cases where the mistake or ambiguity is about the table, column, or value prediction. There is no easy way to resolve structural mistakes or ambiguities if the query sketch is wrong. \mname, on the other hand, offers the potential to recover from such mistakes if the correct query is among the top beam results. This is an orthogonal contribution that could be integrated with other user-interaction modes. Finally, the NaLIR system \cite{Li2014NaLIRAI} has a similar feature allowing the user to pick from multiple interpretations of the input question. However, NaLIR's interpretation is based on syntactical parses of the question rather than interpreting the final semantic parses directly. A rule-based semantic parser then maps the selected syntactic parse to SQL.  As the syntactic parse is not guaranteed to be mapped to the correct SQL, this interpretation does not completely close the gap between what the NLDB performs and what the user thinks it does.

\section{Conclusion}
\vspace{\secvspace}
We presented \mname, a natural language interface to databases (NLDB) that is accurate, interpretable, and works on a wide range of domains. Our system explains its actions in natural language so that the user can select the right answer from multiple hypotheses, capitalizing on the much higher beam accuracy instead of top-1 accuracy. \mname provides a complementary way to resolve mistakes and ambiguities in NLDB.

\section*{Acknowledgments}
We appreciate the ACL demo anonymous reviewers for their valuable inputs. We would like to thank Mehrsa Golestaneh and April Cooper for their work on the improved front-end version, \url{https://turing-app.borealisai.com}, which is not in the scope of this publication. We also would like to thank Wendy Tay and Simon J.D. Prince for their general support.

\bibliography{acl2021}
\bibliographystyle{acl_natbib}

\end{document}